\documentclass[letterpaper, 10 pt, conference]{ieeeconf}

\usepackage[utf8x]{inputenc}
\usepackage{cite}
\usepackage{amssymb,amsfonts,bm}
\usepackage{algorithmic}
\usepackage{graphicx}
\usepackage{textcomp}
\usepackage{amsmath,tikz}
\usepackage{color}
\usepackage{url}
\usepackage{mathtools}
\usepackage[hidelinks]{hyperref} 
\usepackage{rotating}
\usepackage{blkarray}
\usepackage{physics}
\usetikzlibrary{arrows}
\let\proof\relax
\let\endproof\relax
\usepackage{amsthm}

\usepackage{caption}
\usepackage{subcaption}

\usepackage{etoolbox}
\let\classAND\AND
\let\AND\relax
\usepackage{algorithm}
\usepackage{algorithmic}
\let\AND\classAND
\AtBeginEnvironment{algorithmic}{\let\AND\algoAND}

\floatname{algorithm}{Algorithm}

\graphicspath{{eps/}{png/}}
\DeclareSymbolFont{symbolsC}{U}{pxsyc}{m}{n}
\DeclareMathSymbol{\coloneqq}{\mathrel}{symbolsC}{"42}

\newcommand{\vertiii}[1]{{\left\vert\kern-0.25ex\left\vert\kern-0.25ex\left\vert
		#1 
		\right\vert\kern-0.25ex\right\vert\kern-0.25ex\right\vert}}
\usepackage{multirow}
\usepackage{makecell}
\usepackage{adjustbox}
\usepackage{float}

\newcommand*{\vecbf}[1]{\bm{#1}} % vector boldface
\usepackage{pifont}% http://ctan.org/pkg/pifont
\usepackage{soul}

\newcommand{\pos}{p}

\newcommand{\weight}{w}
\newcommand{\cost}{J}
\newcommand{\feas}{f}
\newcommand{\yaw}{\theta}
\newcommand{\yawrate}{{\dot{\theta}}}
\newcommand{\obst}{o}
\newcommand{\dynobst}{h}
\newcommand{\staobst}{s}
\newcommand{\controlin}{{\vecbf{u}}}
\newcommand{\Penalty}{P}
\newcommand{\nominal}[1]{{#1}_\text{th}}
\newcommand{\states}{\vecbf{x}}
\newcommand{\action}{\vecbf{\weight}}
\newcommand{\actionset}{W}
\newcommand{\statetset}{S}
\newcommand{\observationspace}{\Omega}
\newcommand{\observation}{o}

\newcommand{\observationprob}{O}
\newcommand{\cummulativereward}{\mathcal{R}}
\newcommand{\forget}{\gamma}
\newcommand{\networkparam}{\phi}

\newcommand{\transition}{\mathcal{T}}

\IEEEoverridecommandlockouts

\begin{document}
	
	\title{Learning Dynamic Weight Adjustment for Spatial-Temporal Trajectory Planning in Crowd Navigation}
	\author{Muqing~Cao$^\ast$, Xinhang~Xu$^\ast$, Yizhuo~Yang$^\ast$, Jianping Li, Tongxing Jin, Pengfei Wang, \\ Tzu-Yi Hung, Guosheng Lin, and Lihua Xie$^1$~\IEEEmembership{Fellow,~IEEE}% <-this % stops a space
	\thanks{$^\ast$Equal contribution. M. Cao, X. Xu, Y. Yang, J. Li, T. Jin, P. Wang and L. Xie are with School of Electrical and Electronic Engineering, Nanyang Technological University, 50 Nanyang Avenue, Singapore 639798. T.-Y. Hung is with Delta Electronics Inc. }%
   \thanks{$^1$ Corresponding author (elhxie@ntu.edu.sg).}
   \thanks{This research was conducted under project WP5 within the Delta-NTU Corporate Lab with funding support from A*STAR under its IAF-ICP programme (Grant no: I2201E0013) and Delta Electronics Inc.}
   }

% \markboth{IEEE Robotics and Automation Letters. Preprint Version. Accepted Jan, 2022}
% {CAO \MakeLowercase{\textit{et al.}}: } 	
\maketitle
                           
\begin{abstract}
Robot navigation in dense human crowds poses a significant challenge due to the complexity of human behavior in dynamic and obstacle-rich environments.
In this work, we propose a dynamic weight adjustment scheme using a neural network to predict the optimal weights of objectives in an optimization-based motion planner. 
We adopt a spatial-temporal trajectory planner and incorporate diverse objectives to achieve a balance among safety, efficiency, and goal achievement in complex and dynamic environments.
We design the network structure, observation encoding, and reward function to effectively train the policy network using reinforcement learning,
allowing the robot to adapt its behavior in real time based on environmental and pedestrian information. 
Simulation results show improved safety compared to the fixed-weight planner and the state-of-the-art learning-based methods, 
and verify the ability of the learned policy to adaptively adjust the weights based on the observed situations. 
The feasibility of the approach is demonstrated in a navigation task using an autonomous delivery robot across a crowded corridor over a $300$ m distance. Supplementary Video: \href{https://youtu.be/nSCbNaaF_VM}{https://youtu.be/nSCbNaaF\_VM}
\end{abstract}

% \begin{IEEEkeywords}
% 	Constrained Motion Planning, Aerial Systems: Perception and Autonomy, Optimization and Optimal Control.
% \end{IEEEkeywords}

\IEEEpeerreviewmaketitle

\section{Introduction}
\label{sec:intro}

% \begin{figure}[t!]
%     \centering
%     \begin{subfigure}[b]{\linewidth}
%         \includegraphics[width=0.98\linewidth]{fig/longtraj.png}
%     \end{subfigure}
%     \par\smallskip    
%     \centering
%     \begin{subfigure}[b]{\linewidth}
%         \includegraphics[width=0.98\linewidth,trim={0cm 3cm 0 0},clip]{fig/snapshot2.png}
%     \end{subfigure}
%     \caption{(Top) Trajectory generated using our proposed method in a two-room environment, the safe corridor consists of $34$ polyhedra. (Bottom) Snapshots of a quadrotor executing the trajectory generated.}
% 	\label{fig:longtraj}
% \end{figure}
In recent years, new robot applications have emerged that require robots to operate in close proximity to humans, e.g., parcel delivery in living areas, dish serving in restaurants, surveillance in crowded places, etc. 
Robot navigation in dense human crowds is challenging due to the complex human behaviors in a dynamic and interacting environment \cite{Phani2024survey}. 
Although existing works focus on predicting human motion behaviors using the classic model \cite{1995social} or learning-based methods \cite{salzmann2020trajectron,xu2022remember}, they only achieve accurate predictions over a short horizon.
As a result, the robot navigation in the crowd relies on fast and reactive motion planning to ensure safety and social compliance in dynamic humans \cite{Poddar2023crowd}.

Existing methods for robot navigation in crowds can be classified into classic and learning-based methods.
Classic planning methods employ model-based techniques, including velocity obstacle \cite{van2008Reciprocal, Truong2017toward}, graph search \cite{cao2019dynamic}, and model predictive control\cite{Vulcano2022safe,Mavrogiannis2023winding}, to generate trajectories that satisfy the dynamic feasibility and guarantee non-collision under certain assumptions of human motion model.
Many classic planning methods, such as the dynamic window approach (DWA) \cite{Fox1997dynamic} and model predictive control (MPC), are formulated as optimization problems to obtain the optimal trajectory or control inputs that satisfy a weighted combination of objectives, such as safety, efficiency, and goal attainment.
However, choosing the appropriate weights for the application scenarios often requires multiple trials, which is time-consuming.
Recently, learning-based methods have become mainstream in solving the robot navigation in crowd. 
Reinforcement learning is used to train policies that directly map observations of surrounding humans and static obstacles to robot actions such as speed and rotation rate \cite{Chen2019crowd,Everett2021collision,liu2022intention,xie2023drl}.
The deep network learns human-human and human-robot interactions through trial and error and generates safe and socially compliant navigation commands.
Inverse reinforcement learning is also employed to learn a cooperative navigation strategy from human demonstration \cite{Peter2010learning,kretzz2016socially}.
% However, the actions are chosen from discrete action sets, resulting in abrupt and non-smooth commands.
% For example, the robot's velocities may be commanded to change abruptly (the command acceleration is not considered), causing discomfort to the nearby pedestrians.
%This is especially necessary for non-holonomic car-like robots commonly used to deliver bulky items.
To address the non-holonomic kinematic constraints, \cite{Patel2021DWARL} designed a network that chooses from an ordered set of feasible actions.
However, the real-world applications of learning-based approaches are hindered by their generalizability to unseen environments and lack of safety guarantees.

% The well-known social force model, which has proven effective in simulating human crowd dynamics, states that each human is governed by the attraction force towards his goal and the repulsive force from obstacles and other people.
% Our observation is that humans dynamically adjust the balance between these two components to maintain safe and efficient behavior: 
% in a crowded and narrow walkway, pedestrians may tolerate closer inter-person distances than they would in large open spaces; 
% when turning at a corner with limited visibility, humans tend to stay away from the corner to avoid collision with potential incoming humans.

\begin{figure}[]
    \centering
    \includegraphics[width=1.0\linewidth]{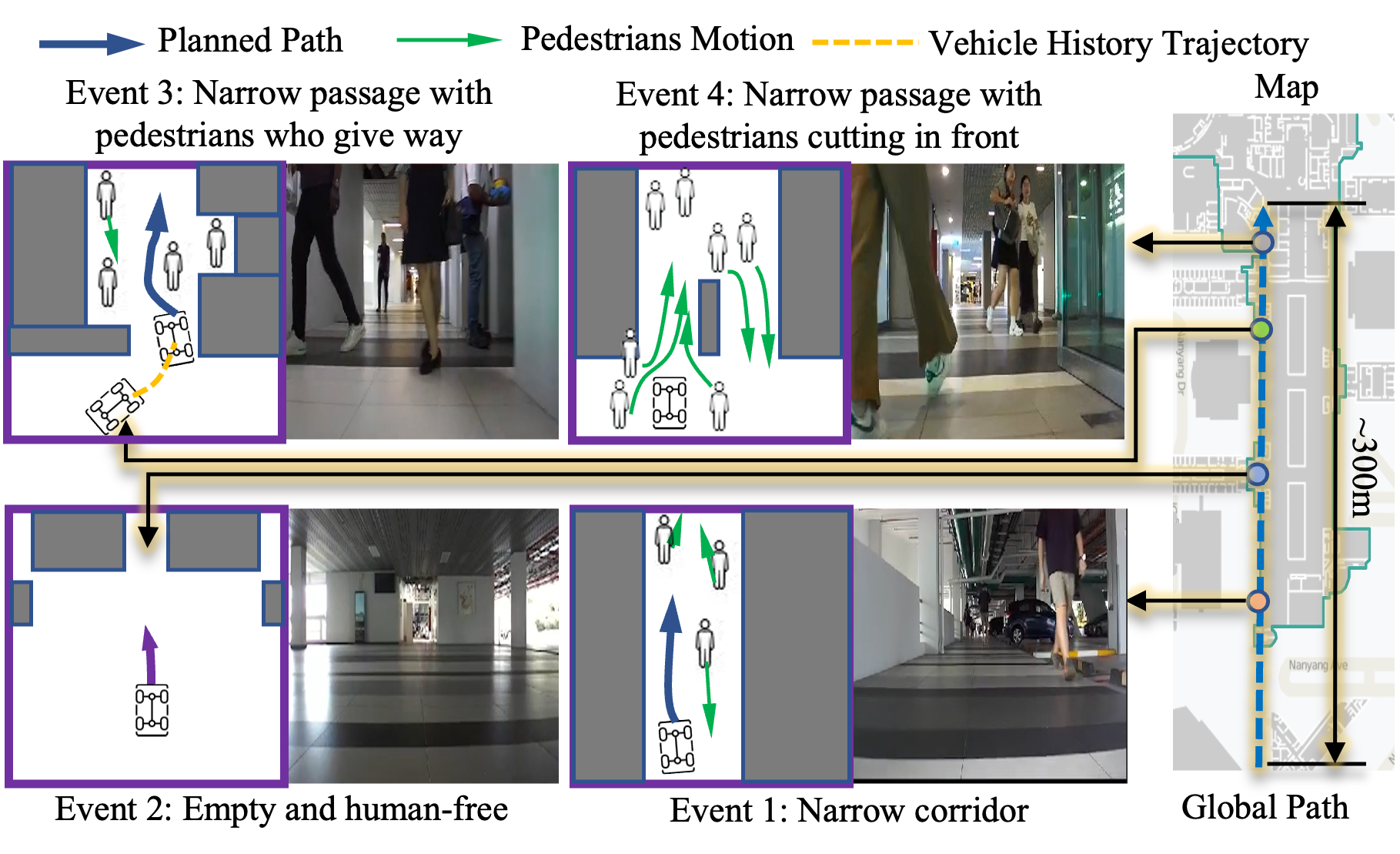}
    \caption{The real-world experiment of the proposed method.}
    \label{fig: real_world}
\end{figure}

To design a robot navigation system capable of effectively maneuvering through human crowds, we draw inspiration from the strategies humans use to navigate such environments, guided by principles from the well-established social force model.
This model explains that individuals are influenced by a combination of attractive forces that pull them toward their desired goals and repulsive forces that push them away from obstacles and other people. 
We observe that humans's movements are not influenced by these components in a fixed manner; rather, they continuously and dynamically adjust the balance of these forces based on the specific context. 
For instance, when walking through a crowded and narrow corridor, people might accept smaller personal distances; 
when approaching a blind corner, individuals tend to slow down and steer slightly wider to avoid unexpected encounters with others from the opposite direction.

Inspired by these adaptive human behaviors, our approach to robot navigation involves dynamically adjusting the weights of various objectives in the robot’s navigation cost function. 
%For example, the robot might increase the weight of repulsive forces to maintain a larger buffer zone from obstacles when there's a risk of sudden appearances by humans, or it might reduce its speed in situations where precise tracking is critical, such as in a densely populated area with unpredictable movements. 
By allowing the robot to modify these weights in real-time based on the surrounding environment, we enable it to navigate more intelligently and responsively, achieving a balance among safety, efficiency, and goal attainment in complex, dynamic crowd scenarios.
We implement this adaptive strategy using a neural network that predicts the optimal weights for each objective in an optimization-based motion planner, considering real-time environmental and pedestrian information. 
The network is trained in a simulated robot navigation task in a realistic environment through reinforcement learning. 
%allowing it to learn effective strategies by interacting with dynamic scenarios. 
Similar concepts have been explored in recent works \cite{Dobrevski2024Dynamic,XIAO2022104132}, where networks are trained to predict the weights for a DWA planner for robot navigation. However, the DWA planner has limitations, as it does not account for the future trajectories of dynamic objects, making it less suitable for navigating in dynamic crowds. Additionally, the DWA cost function only considers a limited set of objectives—such as goal direction, obstacle avoidance, and speed. 
In contrast, our approach utilizes a spatial-temporal trajectory planner that optimizes both the geometric profile and the duration of the trajectory. By expanding the optimization scope to include both spatial and temporal dimensions, and by incorporating a broader range of objectives—such as distances to humans and turn rates—our method generates more diverse motions and behaviors, enhancing the efficiency and safety of robot navigation in congested and compact space.

The main contributions of this work are as follows: \begin{itemize} 
\item We propose a dynamic weight adjustment scheme for spatial-temporal trajectory optimization, considering a diverse range of objectives to enhance robot navigation. 
\item We design the network structure, observation encoding, and reward function to effectively train the policy network using reinforcement learning. 
\item We verify the safety and efficiency of the proposed approach through extensive simulations, comparing it against the state-of-the-art methods. 
\item We demonstrate the real-world applicability of the approach by deploying it on a robot to navigate a long and crowded corridor successfully. 
\end{itemize}

% The contributions are summarized as follows: 
% \begin{itemize}
%     \item We propose a dynamic weight adjustment scheme for spatial-temporal trajectory optimisation considering a diverse set of objectives.
%     \item We design informative observation encoding and reward function to train the policy network using reinforcement learning;
%     \item The safety and efficiency of the proposed approach are verified in several simulation scenes and compared with state-of-the-art approaches;
%     \item We conduct real-world experiments of the approach using a real commercial delivery robot to navigate through a long and crowded corridor.
% \end{itemize}

This work is organized as follows. We first introduce the formulation of a spatial-temporal optimization problem for robot planning in a crowd in Section \ref{sec:spatialtemporal}.
Then, we introduce the framework for learning weight adjustment in Section \ref{sec:learning}, focusing on the design of network and observation encoding.
The training setup is detailed in Section \ref{sec: training}.
In Section \ref{sec: evalua}, we analyze the simulation and experiment results.
Section \ref{sec:conclu} concludes the paper.

% \section{Related Work}

\section{Spatial Temporal Trajectory Optimization}
\label{sec:spatialtemporal}
In this section, we describe the spatial-temporal optimization approach for ground robot navigation and explain our cost function design. A generic spatial-temporal trajectory optimization problem is expressed as:
\begin{equation}
\label{eq:objective}
	\min_{\states(t),T}\cost = \cost_\controlin+ \weight_TT+\Penalty,
\end{equation}
where $\states(t)$ are the robot states at time $t$, including robot position $\vecbf{\pos}(t)$, velocity $\vecbf{v}(t)$, acceleration $\vecbf{a}(t)$, and yaw $\yaw(t)$.
$T$ is the total duration of the trajectory. 
$\cost_\controlin$ is the cost on control effort expressed as 
\begin{equation}
\label{eq:controlin}
\cost_\controlin = \int_{t=0}^{t=T}\lVert\controlin(t)\rVert ^2 dt,
\end{equation}
where $\vecbf{u}(t)$ is the control input. $\weight_TT$ is the time cost weighted by $\weight_T>0$. 
$\Penalty$ includes other additional objectives or penalty terms. Specifically, in our formulation, we adopt the following terms:
% \begin{equation}
% \label{eq:penalty}
% 	\Penalty = \weight_\feas\cost_\feas + \weight_\yawrate\cost_\yawrate + \weight_\obst\cost_\obst + \weight_\dynobst\cost_\dynobst.
% \end{equation}

\subsubsection{Feasibility Cost}
%Soft penalty allows for greater flexibility, enabling the robot to exhibit a wider range of behaviors, especially in dynamic environments where strict adherence to velocity limits might hinder performance or safety. The soft threshold is a way to balance between optimal performance and safety or energy efficiency considerations.

\begin{equation}
\label{eq:feasibility}
\cost_\feas =\int_{t=0}^TL_1(\norm{\vecbf{v}(t)}^2-\nominal{v}^2)+L_1(\norm{\vecbf{a}(t)}^2-\nominal{a}^2)dt.
\end{equation}
Feasibility cost penalizes velocity and acceleration higher than the preset values, $\nominal{v},\nominal{a}$.
$L_1$ is a first-order relaxation function for continuous differentiability. 
Instead of enforcing the dynamic limits as hard constraints in the optimization, we implement a soft penalty 
to allow for more flexibility in the robot's behavior, enabling it to exceed the threshold if necessary while still influencing the optimization process to prefer velocities below this threshold.

\subsubsection{Yaw rate cost}
\begin{equation}
\label{eq:yawrate}
\cost_\yawrate =\int_{t=0}^TL_1(\yawrate(t)^2-\nominal{\yawrate}^2).
\end{equation}
We choose to penalize the yaw rate above a threshold $\nominal{\yawrate}$ because abrupt changes in the robot's heading can cause discomfort or anxiety to nearby humans, forcing them to react quickly to avoid potential collisions. 
Fast heading changes can also cause slippage and affect path-tracking accuracy, increasing the likelihood of collision.

\subsubsection{Static and dynamic obstacle costs}
Static and dynamic obstacle costs,  $\cost_\staobst, \cost_\dynobst$ are expressed similarly:
\begin{equation}
\label{eq:obst}
\cost_\obst =\int_{t=0}^TL_1(d_{o,\text{th}}-d_o(t))dt,\;\,\obst\in\{\staobst,\dynobst\}.
\end{equation}
$d_\staobst(t),d_\dynobst(t)$ are the distances to the closest static and dynamic obstacles (pedestrians), respectively; $d_{\staobst,\text{th}}$ and $d_{\dynobst,\text{th}}$ are the safety clearance thresholds. 

Considering all the objectives mentioned above, the overall optimization problem considered in this work is
\begin{equation}
\label{eq:objective_big}
	\min_{\states(t),T}\cost = \cost_\controlin+ \weight_TT+\weight_\feas\cost_\feas + \weight_\yawrate\cost_\yawrate + \weight_\staobst\cost_\staobst + \weight_\dynobst\cost_\dynobst.
\end{equation}

We adopt the state-of-the-art approach of spatial-temporal trajectory optimization \cite{han2024efficient}, which represents the robot trajectory $\vecbf{\pos}(t)$ as pieces of polynomial curves and solves efficiently using a quasi-Newton method. 
Specifically, we represent trajectories as $5$-th order polynomial curves and consider jerk the control input.
Given the differential flatness of a car-like robot \cite{Murray95differentialflatness}, the robot states required in the above optimization problem can be expressed in terms of robot position $\vecbf{\pos}(t)$ and its derivatives.
The penalty terms are approximated using discretization by sampling robot states evenly along the trajectory.
The geometric profile of a pedestrian is represented as a polygon, and the computation of $d_\dynobst(t)$ follows the approach in \cite{han2024efficient}.

\section{Learning to adjust the weights}
\label{sec:learning}

\begin{figure*}%
\centering
\includegraphics[width=0.9\textwidth]{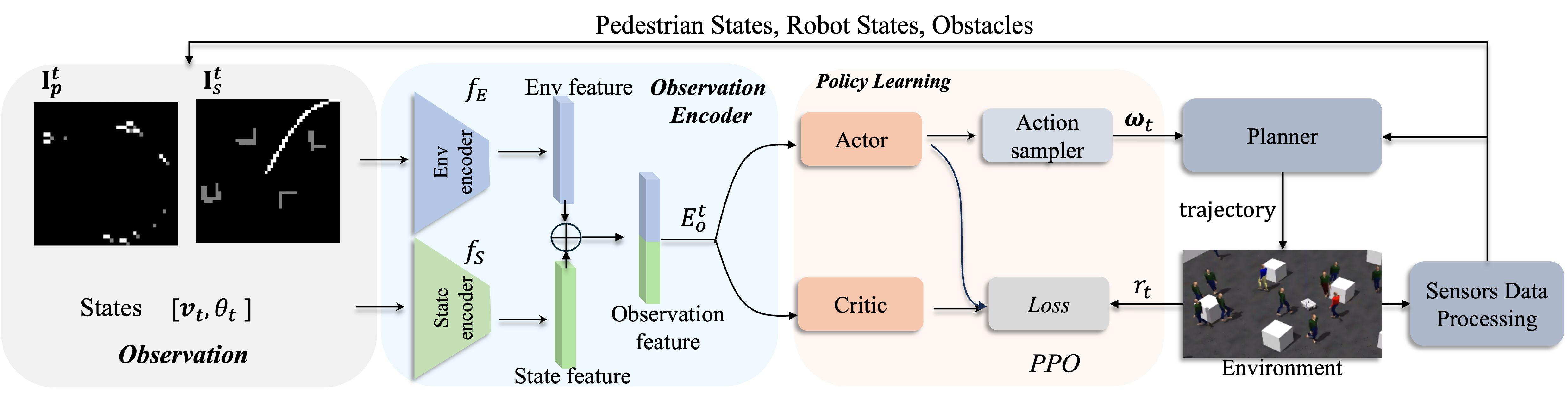}
\caption{Diagram illustrating proposed navigation system that integrates sensor data processing with policy learning to adjust weight of planner. }\label{fig: frame}
\end{figure*}

To dynamically adjust the weights of the optimization problem described in the previous section, we formulate a partially observable Markov decision process (POMDP), denoted as a tuple $(\statetset,\actionset,\transition,R,\observationspace,\observationprob)$, where $\statetset$ denotes the state space, $\actionset$ is the action space, $\transition$ denotes the transition model, $R$ denotes the reward function, $\observationspace$ denotes the observation space, and $\observationprob$ denotes the observation probability model.
The observable states include the robot states $\states(t)$ and the environmental information observed from onboard sensors, including nearby obstacle profile and the human position and velocity.
The unobservable states are the pedestrian's goal location and route preference.
At each step, the robot obtains an observation $\observation_t\in\observationspace$, and the policy $\pi_\networkparam(\cdot|\observation_t)$ outputs an action $\action_t\in\actionset$, which are the weights for the objective function of the spatial-temporal trajectory planner:
\begin{equation}
    \action_t = (\weight_T,\weight_\feas , \weight_\yawrate , \weight_\staobst , \weight_\dynobst).
\end{equation}
Using the updated weights, the trajectory planner generates the future trajectory of the robot by solving the optimization problem \eqref{eq:objective_big}. 
We represent the policy $\pi_\networkparam$ as a neural network to be trained using reinforcement learning, where $\networkparam$ are the network parameters.
%$\action_t\sim\pi(\cdot|\observation_t)$
The objective of the training is to optimize $\networkparam$ to maximize the expected cumulative reward:
\begin{align}
\cummulativereward(\networkparam)=\mathbb{E}_{\pi_\networkparam}\left[\sum_{t=0}^\infty\forget^tr\right],
\end{align}
where $\forget\in[0,1]$ is a discount factor. In the following, we detail design of observation space, network and the reward.

\subsection{Observation Space}\label{sec:observation}
To effectively predict the weights, 
we design the observation space to encompass the vehicle's state, the environmental information, and the goal-related information. 
As shown in the left part of Fig. \ref{fig: frame}, the observation $\observationspace$ is structured into three components: static environment map $\mathbf{I}_s$, dynamic pedestrian map $\mathbf{I}_p$, and the vehicle's kinematic state $\mathbf{X}=\{\boldsymbol{v},\yaw \}$. 
%where $\boldsymbol{v} \in \mathrm{R}^2$ and $\yaw \in [-180,180)\deg$ represent the current velocity and heading of vehicle in global frame. 
$\mathbf{I}_s$ and $\mathbf{I}_p$ are two $50 \cross 50$ 2D grid maps with a resolution of 0.1 meters centered on the robot and axes-aligned with the global frame. 
In $\mathbf{I}_s$, an unoccupied or unknown grid is colored in black, and an occupied one is in grey.
We further encode the planned trajectory from the previous planning cycle into $\mathbf{I}_s$ by coloring the grid containing the planned path as white.
Integrating the previous plan into the local map informs the policy of the desired direction toward the goal without specifying the goal coordinates.
This facilitates improved learning of latent relationships during training without overfitting to fixed locations in particular environments.
Showing the path on the local obstacle map highlights the obstacles near the path, allowing the network to focus on the important obstacles that affect the path shape.

% First, the pre-planned trajectory encodes the outcomes of past weight adjustments of the planner in a latent high-dimensional space, serving as compressed historical data. This compression may encompass pedestrians' reactions to the vehicle as well as the network's adaptability to the environment. Compared to directly stacking past observations, this compressed format significantly reduces the input dimensionality, which helps maintain the completeness of the network’s inputs while simultaneously reducing the parameter count. Second, pre-plan information captures the planner’s behavior in response to specific goal points, inherently embedding goal-point data. Consequently, the network can primarily rely on relative measurements in its observation space, reducing dependency on global positional information. 
% This facilitates improved learning of latent relationships during training without overfitting to fixed locations in particular environments. Furthermore, this reliance on relative measurements enhances the system’s robustness, as the algorithm becomes less sensitive to the precision of upstream algorithms, such as SLAM, thereby supporting plug-and-play functionality in practical applications.

To represent the positions and velocities of varying numbers of pedestrians, the dynamic pedestrian map $\mathbf{I}_p$ features three states: unoccupied (black), currently occupied by humans (gray), and predicted motion of humans (white).
The estimated motion is an approximation based on the assumption of constant pedestrian velocity. 
%Given that the current pedestrian position holds significantly more importance for the system than future motion predictions, 
The encoding process first assigns color to grids representing predicted motion, followed by an overlay of color for current positions to ensure that priority is given to the pedestrian location. 
\vspace{-1 pt}
\subsection{Network Design}
\begin{figure}
    \centering
    \includegraphics[width=1.0\linewidth]{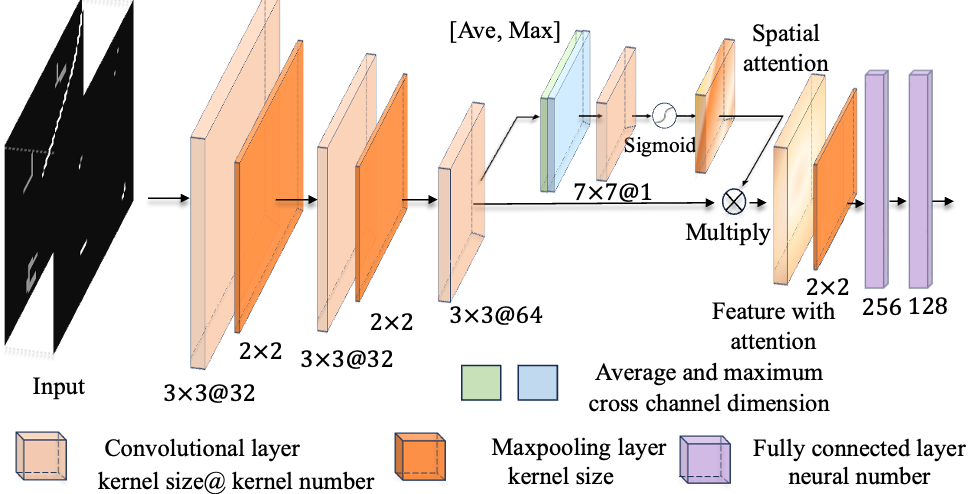}
    \caption{The structure of the environment encoder.}
    \label{fig:env_encoder}
\end{figure}

The overall workflow of the framework is illustrated in Fig. \ref{fig: frame}. At each time step $t$, the observed $\mathbf{I}_s^t$ and $\mathbf{I}_p^t$ are concatenated along the channel dimension. This concatenated representation is then encoded by a CNN-based environment encoder $f_E(.)$ as shown in Fig. \ref{fig:env_encoder}. Specifically, the input map is initially passed through three convolutional layers, each followed by batch normalization, a ReLU activation function and a maxpooling layer. During navigation, regions with dense obstacles or pedestrian and pre-planned trajectories are more crucial for determining optimal paths, while others have negligible influence. To enhance the robot's capability to prioritize these significant regions for effective path planning, a spatial attention mechanism is incorporated into the last convolutional layers. Finally, the feature map with attention is fed into two fully connected layers to obtain the environment embedding.
Concurrently, the agent kinematic state $\mathbf{X}_t$ is input into a MLP-based state encoder $f_S(.)$ which consists of two fully connected layers. The final observation encoding is given by concatenating the environment and state encoding: $E^t_o=[f_E([\mathbf{I}_s^t, \mathbf{I}_p^t]),f_S(\mathbf{X}_t)]$. 
This observation embedding is subsequently used for policy learning. The policy learning network consists of an actor network $A_\pi(E^t_o)$ and a value network $V_\pi(E^t_o)$  to learn the action distribution and the score of the current state. Both the actor and critic networks are based on MLPs, which consist of two fully connected layers with 128 and 64 neurons. Finally, the next action $\action_t$ is selected using an actor sampler from the output action distribution to decide the parameters of the planner.

We utilize Proximal Policy Optimization (PPO) \cite{schulman2017proximal} to train the network. The discount factor $\gamma$ is set to 0.99 to appropriately weigh future rewards. The learning rates for the observation encoder and the critic network are set to $1\cross10^{-3}$, while the actor network is trained with a learning rate of $3 \times 10^{-4}$. In order to balance exploration and exploitation, the standard deviation for the action sampler is initialed at 0.6, with a decay rate of 0.05 for every 200 episodes until reaching a minimum value of 0.1. 
\subsection{Reward}

The reward function is designed to allow the robot to reach the target while minimizing time and collisions.
Specifically, the reward at every time step $t$ is computed as:
\begin{align}
    r_t=r_\text{time}+r_\text{collision}+r_\text{goal}.
\end{align}
The time reward is a dense negative reward to penalize every time step spent:
\begin{align}
    r_\text{time} = -10.
\end{align}
The collision reward is a sparse negative reward to penalize collision with static obstacles or pedestrians. 
In real-world scenarios, 
%people may become curious about autonomous vehicles and approach too closely, leading to close contacts that are not a result of the vehicle’s decision-making. 
the severity of a collision varies with the vehicle’s speed. 
Based on safety tests of the robot model in use, collisions at speeds below $0.4$ m/s pose little damage. 
Therefore, to help the network better learn to avoid catastrophic collisions, we set 
$v_{safe}=0.4$ m/s as the threshold, applying greater penalties to collisions that occur above this threshold. 
%At the same time, to ensure that the vehicle can autonomously avoid collisions caused by external human actions, we also apply some penalties to collisions with lower risks.
\begin{align}
    r_\text{collision} = \begin{cases}
-150, &\text{if in collision and} \|\boldsymbol{v}\|_2  \geq v_{safe}\\
-50, &\text{if in collision and} \|\boldsymbol{v}\|_2  < v_{safe}\\
0, &\text{otherwise.}
\end{cases}
\end{align}
The goal reward is a large positive reward for reaching the goal at the end of the task.
\begin{align}
    r_\text{goal} = \begin{cases}
50, &\text{if goal is reached,}\\
0, &\text{otherwise.}
\end{cases}
\end{align}

\section{Training Setup}\label{sec: training}
\begin{figure}
    \centering
    \includegraphics[width=1\linewidth]{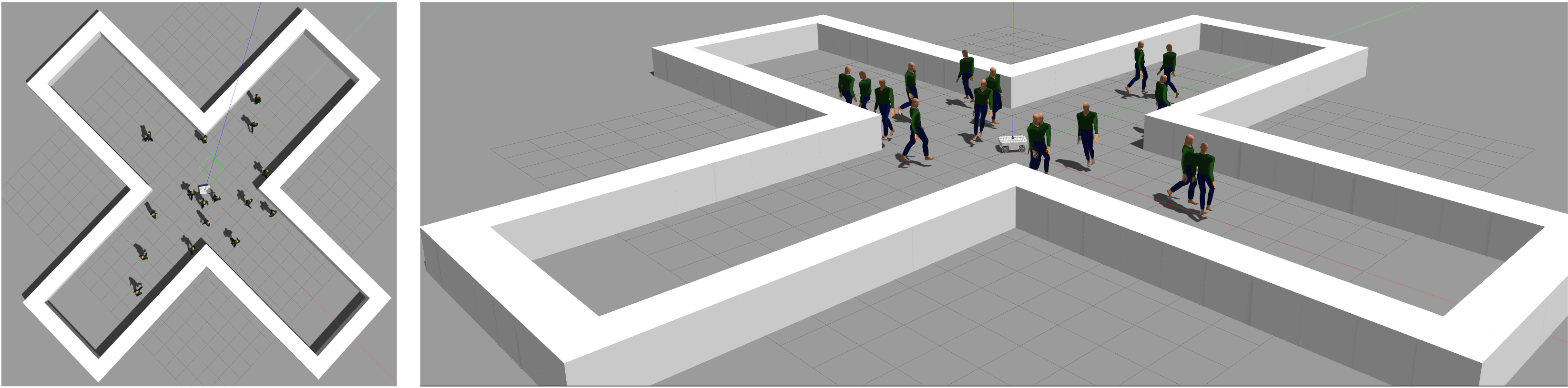}
    \caption{Training environment.}
    \label{fig: train_env}
\end{figure}
We designed a training environment in the Gazebo simulator to replicate a busy indoor setting, as shown in Fig. \ref{fig: train_env}. 
The environment features two corridors intersecting at a junction, with 17 pedestrians walking back and forth between the ends of the corridors. 
The pedestrians' motion is simulated using the social force model. 
In each episode, the robot starts at one of the four corridor endpoints, with the target set at any of the other three endpoints. 
To ensure an accurate simulation of the robot's dynamics and control, the simulated robot's chassis geometry and kinematic properties, such as wheelbase and wheel size, match those of the actual robot used in our experiments. 
The robot gathers environmental information using an onboard 2D LiDAR sensor, which produces distance measurements to the surroundings at $10$ Hz. 
The positions and velocities of pedestrians are obtained from the simulator. 
The policy network generates a prediction of weights at $1$ Hz and triggers a trajectory replan using the new weights.
Since the environmental information is updated more frequently than the weights, a replan is also triggered when the updated environmental information reveals that the previous plan will cause a collision.

We define two early termination conditions to expedite the neural network’s learning process. 
The first occurs when the planning algorithm fails to generate a feasible and safe trajectory for consecutive planning instances, often due to excessive aggressiveness or over-conservativeness of the behavior caused by the weight adjustment. 
In the former case, the vehicle may have collided with an obstacle, making the initial condition infeasible. In the latter, excessive safety prioritization can prevent the planner from finding a feasible solution. 
The second termination condition occurs when the vehicle fails to reach the target within the allotted time limit. 
In the early termination, a penalty of $-1500$ is imposed.

% To expedite the neural network’s learning process, we define two termination conditions. The first occurs when the planning algorithm fails to complete within a valid time frame, often due to excessive aggressiveness or over-conservativeness of the behavior. In the former case, the vehicle may collide with an obstacle due to failure to track the aggressive trajectory or react to a suddenly appeared obstacle. In the latter, excessive safety prioritization can prevent the planner from finding a feasible solution within a reasonable time. 
% The second termination condition is a time-out, which arises when the vehicle fails to reach the target within the allotted time limit, often due to unfeasible weight configurations. In the event of early termination, a significant penalty of -1500 is imposed the neural network to actively avoid the occurrence of the two aforementioned conditions.

\section{Evaluation}\label{sec: evalua}
We train our policy on a computer with a Nvidia 3090 GPU. 
The planner parameters are set as $\nominal{v} = 1.0$ m/s, $\nominal{a} = 1.0$ m/s$^2$, $\nominal{\yawrate}=0.2$ rad/s,  $d_{\staobst,\text{th}} = 1.0$ m, $d_{\dynobst,\text{th}} = 1.0$ m.
\begin{figure}
    \centering
    \includegraphics[width=1.0\linewidth]{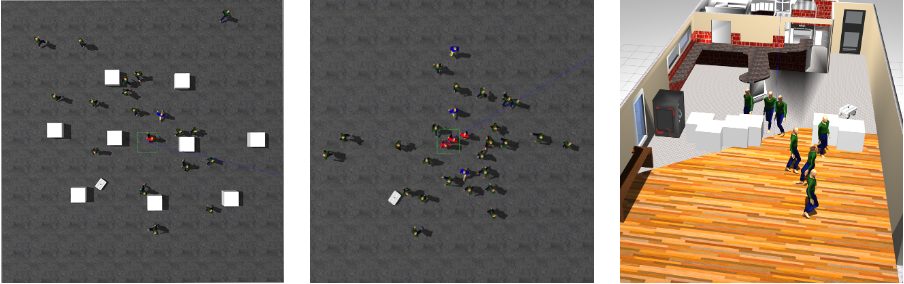}
    \caption{Test scenes from left to right: (1) obstacle- and human-populated scene, (2) obstacle-free and human-dense scene, and (3) narrow indoor scene.}
    \label{fig:test_env}
\end{figure}

\subsection{Simulation}

We evaluate the proposed approach in three challenging test scenes (Fig. \ref{fig:test_env}): (1) Obstacle- and human-populated: an open environment with 9 square obstacles and 19 pedestrians. (2) Obstacle-free and human-dense: an open environment with $31$ pedestrians. (3) An indoor environment with $7$ pedestrians and a narrow entrance.
In each scenario, the robot is tasked with traversing from one end of the environment to the other, where the path crosses regions densely populated with moving pedestrians or static obstacles. 
We compare our proposed approach with the following methods:
\begin{itemize} 
\item Spatial-temporal trajectory planner with fixed weights (ST): the robot uses fixed weights $\vecbf{\weight}$ throughout each test. We evaluate $6$ different sets of weights: one set where all weights are equal to one, and five others where, in each case, one specific weight is set to five.

\item Dynamic Adaptive DWA (DADWA): Dynamic weight adjustment for DWA planner. Since the implementation of in \cite{Dobrevski2024Dynamic} is not publicly available, we trained our own version of DADWA using the same network and observation design as our approach. An open-source DWA planner is used for this purpose\footnote{https://github.com/amslabtech/dwa\_planner}.
\item DRL\_VO \cite{xie2023drl}: A recent learning-based robot planner. We retrain the policy using our robot model with the updated velocity limits. 
\end{itemize}
In each scene, the navigation task is conducted for $100$ runs. The following metrics are recorded for evaluation: (1) mission completeness: a mission is complete if the robot reaches the goal within $60$ seconds; (2) Average time taken and distance traveled by the robot to reach the goal, considering only complete runs; 
(3) collided runs: number of runs in which active collision occurs, and (4) total collision counts (TCC): total active collision events detected. Collision detection is conducted at $50$ Hz using Gazebo contact checking; a single collision count represents an instance where the robot is in contact with its surroundings. 
We only consider active collision where the robot's speed is above $v_\text{safe}$ during the contact,  
because it indicates a dangerous situation where the robot may injure a pedestrian. 

% To better demonstrate the performance differences between the algorithms, we selected the following metrics: Suc. (Success trials): the number of trials the vehicle reaches the goal point out of 100 trials; Avg. Time: the average time required to successfully reach the goal point; \begin{tabular}[c]{@{}l@{}}Avg. \\ Dist\end{tabular}: the average path length traveled to reach the goal; CT (collision trials): the number of trials where active collisions occurred within 100 tests; SAC (sum of active collisions): the total number of active collisions observed in 100 tests. 

\begin{table*}[ht]
\centering
\setlength{\tabcolsep}{4pt}
\renewcommand{\arraystretch}{1.2}
\begin{tabular}{|c|c|c|c|c|c|c|c|c|c|c|c|c|c|c|c|c|c|c|c|c|c|}
\hline
\multirow{2}{*}{Method} & \multicolumn{5}{c|}{Scene 1: obstacle- and human-populated} & \multicolumn{5}{c|}{Scene 2: human-dense} & \multicolumn{5}{c|}{Scene 3: narrow indoor} \\ \cline{2-16} 
 & Complete & \begin{tabular}[c]{@{}l@{}}Avg. \\ Time\end{tabular} & \begin{tabular}[c]{@{}l@{}}Avg. \\ Dist\end{tabular} & \begin{tabular}[c]{@{}l@{}}Colli. \\ Runs\end{tabular} & TCC & Complete & \begin{tabular}[c]{@{}l@{}}Avg. \\ Time\end{tabular} & \begin{tabular}[c]{@{}l@{}}Avg. \\ Dist\end{tabular} & \begin{tabular}[c]{@{}l@{}}Colli. \\ Runs\end{tabular} & TCC & Complete & \begin{tabular}[c]{@{}l@{}}Avg. \\ Time\end{tabular} & \begin{tabular}[c]{@{}l@{}}Avg. \\ Dist\end{tabular} & \begin{tabular}[c]{@{}l@{}}Colli. \\ Runs\end{tabular} & TCC \\ \hline
Proposed & $\underline{98}$ & $\underline{20.2}$ & $\underline{17.9}$ & $\underline{8}$ & $\mathbf{81}$ & $95$ & $42.4$ & $\underline{16.9}$ & $\mathbf{7}$ & $\mathbf{40}$ & $\mathbf{98}$ & $19.4$ & $\underline{17.9}$ & $\mathbf{1}$ & $\mathbf{5}$ \\ \hline
ST(all$=1$) & $93$ & $20.5$ & $18.8$ & $11$ & $328$ & $92$ & $42.7$ & $17.3$ & $\underline{11}$ & $89$ & $90$ & $\mathbf{18.4}$ & $18.1$ & $6$ & $126$ \\ \hline
ST($\weight_{T}=5$) & $94$ & $\mathbf{16.5}$ & $17.9$ & $18$ & $229$ & $85$ & $42.3$ & $17.3$ & $\underline{11}$ & $87$ & $85$ & $\underline{18.8}$ & $18.0$ & $\underline{4}$ & $\underline{98}$ \\ \hline
ST($\weight_{\feas}=5$) & $96$ & $23.6$ & $18.5$ & $\mathbf{7}$ & $\underline{148}$ & $73$ & $\underline{23.7}$ & $18.4$ & $67$ & $3017$ & $27$ & $20.7$ & $19.1$ & $73$ & $3083$ \\ \hline
ST($\weight_{\yawrate}=5$) & $83$ & $21.6$ & $\underline{17.8}$ & $21$ & $300$ & $94$ & $40.1$ & $18.2$ & $20$ & $191$ & $63$ & $22.8$ & $19.1$ & $40$ & $1249$ \\ \hline
ST($\weight_{\staobst}=5$) & $55$ & $29.8$ & $22.6$ & $56$ & $1974$ & $91$ & $40.4$ & $17.3$ & $15$ & $112$ & $69$ & $19.6$ & $\mathbf{17.9}$ & $18$ & $384$ \\ \hline
ST($\weight_{\dynobst}=5$) & $79$ & $23.4$ & $19.9$ & $32$ & $1133$ & $90$ & $44.5$ & $17.3$ & $\underline{11}$ & $\underline{65}$ & $78$ & $18.9$ & $18.2$ & $23$ & $1196$ \\ \hline
DADWA & $\mathbf{99}$ & $20.5$ & $\mathbf{17.7}$ & $39$ & $782$ & $\mathbf{100}$ & $\mathbf{21.3}$ & $\mathbf{16.3}$ & $91$ & $3104$ & $\underline{91}$ & $27.3$ & $21.9$ & $54$ & $2198$ \\ \hline
DRL\_VO & $84$ & $28.8$ & $18.5$ & $43$ & $363$ & $\underline{99}$ & $26.4$ & $21.5$ & $77$ & $1897$ & $69$& $36.1$ &$19.3$ & $43$ & $409$ \\ \hline

\end{tabular}
\caption{Performance results. The best performance for each column is in bold and the second best is underlined.}
\label{Tab: simu_experiment}
\end{table*}

Table \ref{Tab: simu_experiment} shows the simulation result.
We observe that the proposed approach has the most comprehensive performance, achieving high mission completeness and low collision counts across all test scenes.
Among the weights chosen for fixed-weight planning, the balanced configuration ST(all $=1$) achieves the most well-rounded result.
However, compared to our approach, it consistently has lower completeness rates and higher collision runs and counts, indicating that the learned weight adjustment policy is effective in adapting to the environment setting and improving safety.
Compared to the balanced weight setting, we observe that weight settings with a clear focus on particular objectives may result in drastically different performances across scenes.
Specifically, a high weight on human avoidance ($\weight=5$) yields good completeness ($90\%$) and low collision cases ($11$) in obstacle-free human-dense scenes but causes low completeness rates in the other obstacle-ridden scenes ($79\%$ and $78\%$).
In environments with both obstacles and pedestrians, 
a focus on either static or dynamic obstacle avoidance results in frequent failures to reach the target because the robot generates aggressive motion and sharp turns for collision avoidance, which significantly exceed the velocity and acceleration thresholds.
Such aggressive trajectory causes large tracking errors and eventual collision with the obstacles.
On the other hand, putting a high weight on kinematic feasibility ($\omega_f=5$) allows the robot to navigate in obstacle- and human-populated environments safely, indicating that maintaining a good tracking performance is important.
However, in the human-dense region and narrow passage, the robot becomes easily stuck due to surrounding pedestrians and fails to escape the situation due to low speed.
In essence, choosing an appropriate weight combination could benefit navigation in some particular scenarios but our learned policy enables good performance in diverse environments.

DADWA achieves the highest task completeness rate and shortest path length for Scenes 1 and 2; however, it results in active collisions in $91\%$ of runs in the human-dense scene, the highest among all approaches.
In many situations, DADWA generates turning trajectories to avoid the closest pedestrians but hit another nearby pedestrian.
Clearly, the DWA planner with only current sensor information cannot generate safe motion in a dynamic and complex environment.
DRL\_VO also performs unsatisfactorily, with high percentages of collision in all test scenes.

\begin{figure*}
    \centering
    \includegraphics[width=1.0\linewidth]{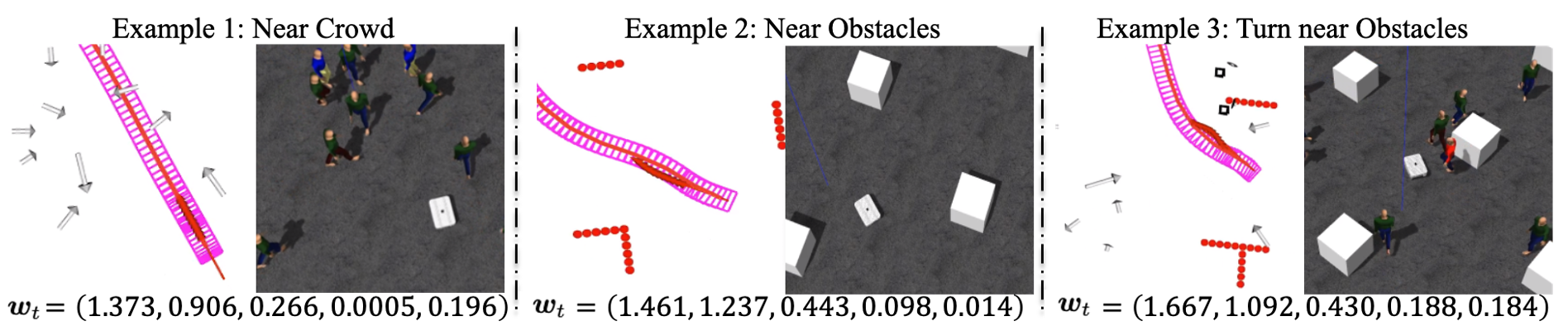}
    \caption{Three scenarios and the policy output $\vecbf{\weight}_t$. The red circles are the detected obstacles, the purple rectangles outline the planned robot footprint, and the gray arrows represent the pedestrians with estimated velocity.}
    \label{fig:weights_adjust}
\end{figure*}

% \subsection{Qualititative Analysis of Policy}
To further analyze the learned policy,  
we examine the policy output $\vecbf{\weight}_t$ in several different situations faced by the robot in the simulation, as shown in Figure \ref{fig:weights_adjust}. 
We observe that the values assigned to each weight are not of the same magnitude:
$\weight_\staobst$ and $\weight_\dynobst$ are always much smaller than $\weight_T$ and $\weight_\feas$.
Therefore, one reason for the superior performance of the approach is that the network learns the appropriate value range of each weight.
Furthermore, it is evident that the weights are adjusted based on the observed situation.
In the human-dense scenario (example 1), the weight for dynamic avoidance $\weight_\dynobst$ is more than $10$ times higher than that in the human-free scenario (example 2). In contrast, the weight for static avoidance $\weight_\staobst$ is almost negligible in example 1 compared to example 2.
In example 3, where both obstacle and humans are present, $\weight_\dynobst$ and $\weight_\staobst$ are set to high values.
In examples 2 and 3, where the robot needs to make turns at the obstacle corner, the weight for the yaw rate is set to higher values to avoid aggressive heading changes and hence ensure reliable tracking.

\subsection{Real-World Experiment}
The proposed approach is implemented and tested using a ground robot for indoor navigation across a long (approximately 300 m) and narrow corridor, as depicted in Fig. \ref{fig: real_world}.
% \textcolor{blue}{Although the inference speed of the policy network can reach 50 Hz on Nvidia Orin-NX, it is limited by the pedestrian estimator’s 2 Hz estimation frequency. Additionally, considering that re-planning should not be overly frequent, we adjust the planner’s weights at a speed of 1 Hz.}
The robot carries two computing devices: the policy inference runs on an Orin NX at $1$ Hz, while the trajectory planner is run on an NUC i7-1260P.
Two computers communicate using the ROS messages through a local network.
Two 2D lidars are mounted on the robot for obstacle and human detection and tracking; human position and velocity are estimated by detecting leg movements through consecutive LiDAR scans\footnote{we use the package https://github.com/wg-perception/people}.
To reach a user-defined goal point, a global path is first generated based on the prior map, and then the proposed approach is used to reach the sub-goals along the global path. 
To ensure safety during the test, we reduce the speed threshold to $\nominal{v}=0.6$.
% An experiment is conducted where the robot travel 
% This corridor serves as a connection between two primary teaching buildings. The experiment was conducted during the dinner period, when the corridor experienced significant foot traffic, resulting in a relatively crowded environment.

We identified four representative events along the path to validate the effectiveness of our algorithm. 
% In Event 1, a pedestrian was advancing at a speed comparable to that of the vehicle. As they neared a narrow gate, the policy network adjusted the controller’s weight $\action (1.749,1.195,0.187,0.0427,0.090)$, providing the planner with sufficient steering flexibility to prevent the pedestrian from making sudden movements at the narrow passage, which could potentially force the vehicle into an unsafe situation.
% In Event 2, a large group of pedestrians appeared in the robot's intended path. In response, the policy network modified the planner's weight configuration to setting $\action (0.645,0.243,1.310,0.096,1.552)$, increasing the attention paid to both static and dynamic obstacles, while simultaneously reducing the emphasis on time optimization. Furthermore, a high weight was assigned to kinematic feasibility to prioritize safety, as previously discussed in the simulation.
Events 1 and 2 demonstrate the ability of the proposed workflow to handle narrow passages and empty environments. In narrow environments with pedestrians, the policy adopts a conservative weight on time. Conversely, upon detecting an open, pedestrian-free environment, the policy increases the weight on time cost, incentivizing a faster trajectory.

Event 3 occurred as the vehicle encountered a crowd after traversing a narrow passage. Before making the turn, the vehicle was unable to detect the pedestrian due to the obstruction. 
Consequently, the policy adjust the planner as $\action_t=(1.897, 1.121, 0.122, 0.446, 0.0971)$, which prioritized static obstacle detection to ensure safety during wide turns while adhering to speed and turning angle constraints. 
Once the vehicle completed the turn and detected the pedestrians, who were waiting, the policy adjusted the weight to $\action_t=(1.591, 0.870, 0.0129, 0.309,0.0993)$. This adjustment reduced $\action_{\dot{\theta}}$ allowing the vehicle to maneuver more flexibly through the crowd. %Simultaneously, the reduction of $\action_{s}$ and a small $\action_{h}$
 %enabled the planner to compute a path in such a crowded environment. 
 In scenarios where the environment is narrow and pedestrians are waiting for the vehicle to pass, this represents an effective strategy.
% Upon initial meet, the policy adjusted the planner’s weight to 
% $\action_t=(1.897,1.121,0.1218,0.446,0.097)$, which balanced the consideration of static and dynamic obstacles. Additionally, the policy increased the priority of steering control and kinematic feasibility to prevent the vehicle from engaging in aggressive maneuvers that could pose a risk to pedestrians or the vehicle itself. Subsequently, upon detecting that the crowd was waiting for the vehicle to pass, the policy adjusted the planner's weight configuration to $\action_t=(1.593, 0.910, 0.308, 0.0129, 0.0976)$ in order to reduce the planner’s sensitivity to environmental changes and rapidly identify a path, thereby preventing a deadlock situation between the vehicle and pedestrians.

Event 4 is similar to Event 3, with the key difference being that, in this case, the pedestrians did not yield to the vehicle but instead opted to cross its path. The policy can adjust the vehicle’s actions accordingly, initiating deceleration and stopping to give way. Once safety was confirmed, the vehicle resumed motion, avoiding the obstruction of the passage. 

\section{Conclusion}\label{sec:conclu}
We introduced a learning-based dynamic weight adjustment scheme for robot navigation in crowded environments. Our approach demonstrated the ability to learn effective strategies for balancing various objectives across different scenarios, leading to comprehensive performance improvements. In the future, we plan to design more diverse simulation environments to further validate the consistency of the learned strategies. The real-world experiment serves as a promising start to assess the reliability of the approach in robot delivery tasks within human-dense areas.

 % \section*{Acknowledgments}
 % This research was conducted under project WP5 within the Delta-NTU Corporate Lab with funding support from A*STAR under its IAF-ICP programme (Grant no: I2201E0013) and Delta Electronics Inc.
 
\bibliographystyle{ieeetr}        % Include this if you use bibtex 
\bibliography{main}

\begin{thebibliography}{10}

\bibitem{Phani2024survey}
P.~T. Singamaneni, P.~Bachiller-Burgos, L.~J. Manso, A.~Garrell, A.~Sanfeliu, A.~Spalanzani, and R.~Alami, ``A survey on socially aware robot navigation: Taxonomy and future challenges,'' {\em The International Journal of Robotics Research}, vol.~0, no.~0, p.~02783649241230562, 0.

\bibitem{1995social}
D.~Helbing and P.~Moln\'ar, ``Social force model for pedestrian dynamics,'' {\em Phys. Rev. E}, vol.~51, pp.~4282--4286, May 1995.

\bibitem{salzmann2020trajectron}
T.~Salzmann, B.~Ivanovic, P.~Chakravarty, and M.~Pavone, ``Trajectron++: Dynamically-feasible trajectory forecasting with heterogeneous data,'' in {\em Computer Vision--ECCV 2020: 16th European Conference, Glasgow, UK, August 23--28, 2020, Proceedings, Part XVIII 16}, pp.~683--700, Springer, 2020.

\bibitem{xu2022remember}
C.~Xu, W.~Mao, W.~Zhang, and S.~Chen, ``Remember intentions: Retrospective-memory-based trajectory prediction,'' in {\em Proceedings of the IEEE/CVF Conference on Computer Vision and Pattern Recognition}, pp.~6488--6497, 2022.

\bibitem{Poddar2023crowd}
S.~Poddar, C.~Mavrogiannis, and S.~S. Srinivasa, ``From crowd motion prediction to robot navigation in crowds,'' in {\em 2023 IEEE/RSJ International Conference on Intelligent Robots and Systems (IROS)}, pp.~6765--6772, 2023.

\bibitem{van2008Reciprocal}
J.~van~den Berg, M.~Lin, and D.~Manocha, ``Reciprocal velocity obstacles for real-time multi-agent navigation,'' in {\em 2008 IEEE International Conference on Robotics and Automation}, pp.~1928--1935, 2008.

\bibitem{Truong2017toward}
X.-T. Truong and T.~D. Ngo, ``Toward socially aware robot navigation in dynamic and crowded environments: A proactive social motion model,'' {\em IEEE Transactions on Automation Science and Engineering}, vol.~14, no.~4, pp.~1743--1760, 2017.

\bibitem{cao2019dynamic}
C.~Cao, P.~Trautman, and S.~Iba, ``Dynamic channel: A planning framework for crowd navigation,'' in {\em 2019 International Conference on Robotics and Automation (ICRA)}, pp.~5551--5557, 2019.

\bibitem{Vulcano2022safe}
V.~Vulcano, S.~G. Tarantos, P.~Ferrari, and G.~Oriolo, ``Safe robot navigation in a crowd combining nmpc and control barrier functions,'' in {\em 2022 IEEE 61st Conference on Decision and Control (CDC)}, pp.~3321--3328, 2022.

\bibitem{Mavrogiannis2023winding}
C.~Mavrogiannis, K.~Balasubramanian, S.~Poddar, A.~Gandra, and S.~S. Srinivasa, ``Winding through: Crowd navigation via topological invariance,'' {\em IEEE Robotics and Automation Letters}, vol.~8, no.~1, pp.~121--128, 2023.

\bibitem{Fox1997dynamic}
D.~Fox, W.~Burgard, and S.~Thrun, ``The dynamic window approach to collision avoidance,'' {\em IEEE Robotics \& Automation Magazine}, vol.~4, no.~1, pp.~23--33, 1997.

\bibitem{Chen2019crowd}
C.~Chen, Y.~Liu, S.~Kreiss, and A.~Alahi, ``Crowd-robot interaction: Crowd-aware robot navigation with attention-based deep reinforcement learning,'' in {\em 2019 International Conference on Robotics and Automation (ICRA)}, pp.~6015--6022, 2019.

\bibitem{Everett2021collision}
M.~Everett, Y.~F. Chen, and J.~P. How, ``Collision avoidance in pedestrian-rich environments with deep reinforcement learning,'' {\em IEEE Access}, vol.~9, pp.~10357--10377, 2021.

\bibitem{liu2022intention}
S.~Liu, P.~Chang, Z.~Huang, N.~Chakraborty, K.~Hong, W.~Liang, D.~Livingston~McPherson, J.~Geng, and K.~Driggs-Campbell, ``Intention aware robot crowd navigation with attention-based interaction graph,'' in {\em IEEE International Conference on Robotics and Automation (ICRA)}, pp.~12015--12021, 2023.

\bibitem{xie2023drl}
Z.~Xie and P.~Dames, ``Drl-vo: Learning to navigate through crowded dynamic scenes using velocity obstacles,'' {\em IEEE Transactions on Robotics}, vol.~39, no.~4, pp.~2700--2719, 2023.

\bibitem{Peter2010learning}
P.~Henry, C.~Vollmer, B.~Ferris, and D.~Fox, ``Learning to navigate through crowded environments,'' in {\em 2010 IEEE International Conference on Robotics and Automation}, pp.~981--986, 2010.

\bibitem{kretzz2016socially}
H.~Kretzschmar, M.~Spies, C.~Sprunk, and W.~Burgard, ``Socially compliant mobile robot navigation via inverse reinforcement learning,'' {\em The International Journal of Robotics Research}, vol.~35, no.~11, pp.~1289--1307, 2016.

\bibitem{Patel2021DWARL}
U.~Patel, N.~K.~S. Kumar, A.~J. Sathyamoorthy, and D.~Manocha, ``Dwa-rl: Dynamically feasible deep reinforcement learning policy for robot navigation among mobile obstacles,'' in {\em 2021 IEEE International Conference on Robotics and Automation (ICRA)}, pp.~6057--6063, 2021.

\bibitem{Dobrevski2024Dynamic}
M.~Dobrevski and D.~Skočaj, ``Dynamic adaptive dynamic window approach,'' {\em IEEE Transactions on Robotics}, vol.~40, pp.~3068--3081, 2024.

\bibitem{XIAO2022104132}
X.~Xiao, Z.~Wang, Z.~Xu, B.~Liu, G.~Warnell, G.~Dhamankar, A.~Nair, and P.~Stone, ``Appl: Adaptive planner parameter learning,'' {\em Robotics and Autonomous Systems}, vol.~154, p.~104132, 2022.

\bibitem{han2024efficient}
Z.~Han, Y.~Wu, T.~Li, L.~Zhang, L.~Pei, L.~Xu, C.~Li, C.~Ma, C.~Xu, S.~Shen, and F.~Gao, ``An efficient spatial-temporal trajectory planner for autonomous vehicles in unstructured environments,'' {\em IEEE Transactions on Intelligent Transportation Systems}, vol.~25, no.~2, pp.~1797--1814, 2024.

\bibitem{Murray95differentialflatness}
R.~M. Murray, M.~Rathinam, and W.~Sluis, ``Differential flatness of mechanical control systems: A catalog of prototype systems,'' in {\em Proceedings of the 1995 ASME International Congress and Exposition}, 1995.

\bibitem{schulman2017proximal}
J.~Schulman, F.~Wolski, P.~Dhariwal, A.~Radford, and O.~Klimov, ``Proximal policy optimization algorithms,'' {\em arXiv preprint arXiv:1707.06347}, 2017.

\end{thebibliography}

\end{document}